\documentclass[conference]{IEEEtran}
\usepackage{times}

\usepackage[numbers]{natbib}
\usepackage{mathtools,amsmath}
\usepackage{multicol}
\usepackage[bookmarks=true]{hyperref}
\usepackage{subcaption}
\usepackage{balance}

\usepackage{graphicx} 
\graphicspath{{figures/}{figures/streched/}}

\newcommand{\argmin}{\arg\!\min}

\begin{document}

\title{Learning Arbitration for Shared Autonomy \\
by Hindsight Data Aggregation}

\author{Yoojin Oh$^{1}$, Marc Toussaint$^{1}$ and Jim Mainprice$^{1,2}$\\
	\vspace{0.1cm}
	\authorblockA{\tt{\small{firstname.lastname@ipvs.uni-stuttgart.de}}}
	\authorblockA{$^1$Machine Learning and Robotics Lab, University of Stuttgart, Germany}
	\authorblockA{$^2$Max Planck Institute for Intelligent Systems ;  MPI-IS ; T{\"u}bingen, Germany}
	\vspace{-0.8cm}
}



%

\maketitle

\begin{abstract}

In this paper we present a framework
for the teleoperation of pick-and-place tasks.
We define a \textit{shared control} policy that
allows to blend between direct user control
and autonomous control based 
on user intent inference.
One of the main challenges in shared autonomy systems is to define the 
\textit{arbitration} function,
which decides when to let the autonomous agent take over.
In this work, we propose a model and training method
to learn the arbitration function.
Our model is based on a recurrent neural network
that takes as input the state, intent prediction scores
and user command to produce an arbitration between user and robot commands.
This work extends our previous work
\let\thefootnote\relax\footnotetext{
Workshop on
``AI and Its Alternatives in Assistive and Collaborative Robotics" (RSS 2019), Robotics: Science and Systems Freiburg, Germany.}
on differentiable policies for shared autonomy \cite{Oh:2019}.
Differentiability of the policy is
desirable to further train the shared autonomy system end-to-end. 
In this work we propose training of the arbitration function
by using data from user performing the task with shared control.  
We present initial results by teleoperating a gripper in a virtual environment
using pre-trained motion generation and intent prediction.
We compare our data aggregation training procedure to a hand-crafted arbitration function.
Our preliminary results show the efficacy of the approach and
shed light on limitations that we believe
demonstrate the need for user adaptability in shared autonomy systems.
\end{abstract}

\IEEEpeerreviewmaketitle

\section{Introduction}

It is anticipated that robots will be able to assist humans in various tasks, 
from assisting people with motored impairments to performing tasks in environments inaccessible
or too dangerous for humans, such as disaster environments, underwater, or outer space. 
However, current robotic technology is unable to achieve full autonomy
and robots still lack in robustness in dynamic environments which makes teleoperation appealing.
Traditional approaches to robot teleoperation \cite{phillips2016autonomy}
rely on the user assigning low-level (\textit{direct control}) or mid-level (\textit{traded control})
commands to be performed by the robot.

Typically, difficulties originate from limited situation awareness
and discrepancy between the human and robot morphologies.
This can lead to human operational errors,
which can be compensated by extensive prior practice.
One way to mediate these difficulties is \textit{shared control},
combining human intelligence with the robot's autonomy. 

Shared autonomy is an active field of research 
\cite{dragan2013policy, Nikolaidis:2017hfa, reddy2018shared,
	Javdani:2018bt}.
These systems rely on two components:
\begin{enumerate}
\item \textbf{prediction} of user intent
\item  \textbf{blending} user vs. autonomous controls.
\end{enumerate}

Arbitration refers to the blending strategy \cite{dragan2013policy}.
One common form of blending is through a linear combination between user and agent policies
\cite{dragan2013policy, allaban2018blended, schultz2017goal, gopinath2016human}.
The arbitration parameter $\alpha$ can depend on different factors such as the confidence in the user intent prediction
\cite{dragan2013policy, gopinath2016human, schultz2017goal},
or considering the difference between each commands
\cite{allaban2018blended}.

\begin{figure}[t]
  \centering
  \includegraphics[clip,trim={1.5cm 1.5cm 1.5cm .5cm},scale=0.3]{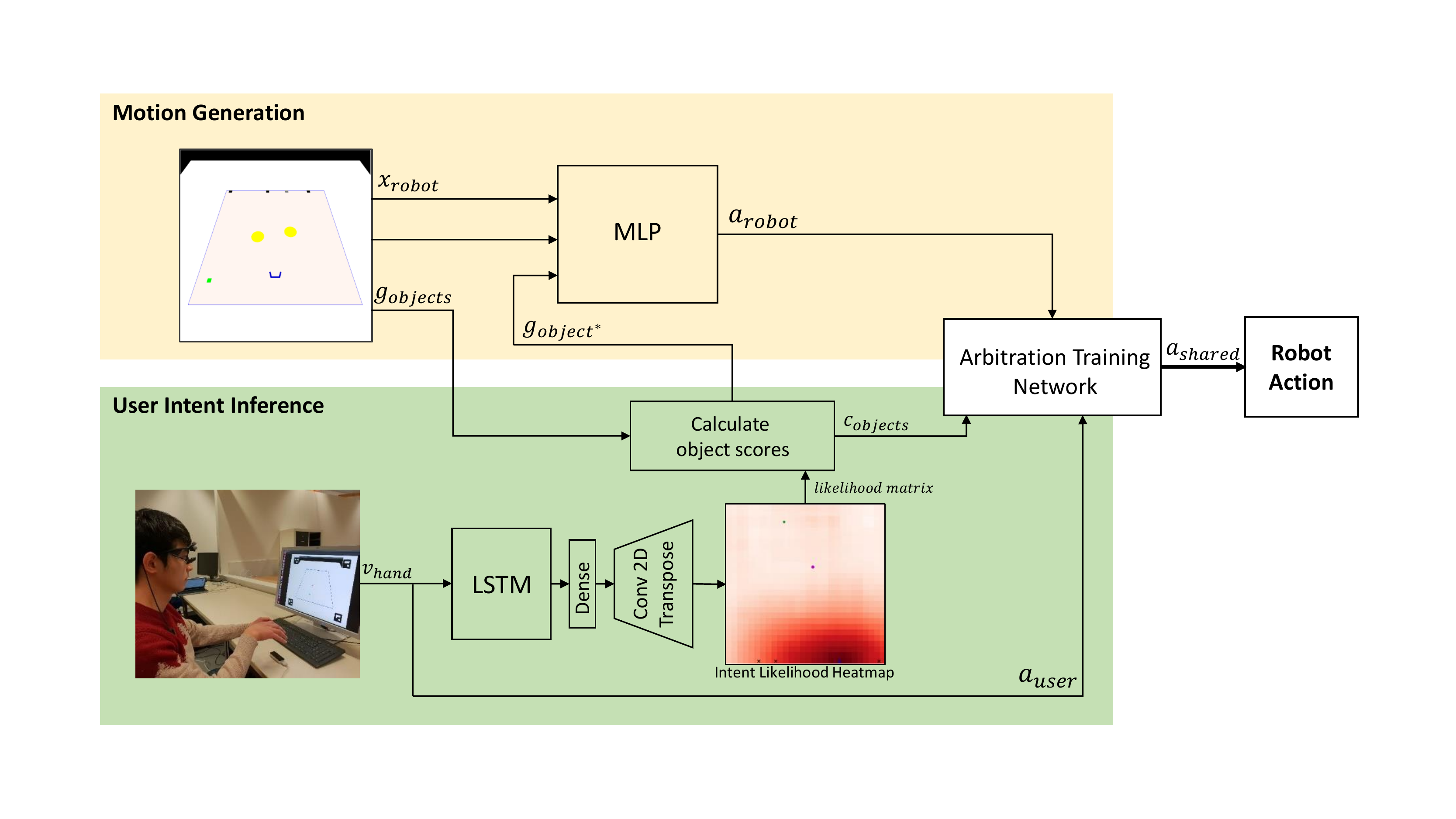}
  \caption{
  Policy Architecture. Virtual environment with 4 target objects (top left). User interacting with the setup (bottom left). 
  The arbitration module blends between autonomous motion generation and direct control of the human. }
  \label{fig:architecture}
  \vspace{-0.7cm}
\end{figure}

In previous work \cite{Javdani:2018bt}, authors have reported that
users prefer to keep control as much as possible and prefer
arbitration over hindsight optimization, despite optimality.
However, defining an arbitration function that is nor too 
\textit{timid} (i.e., only gives assistance when very confident)
nor to \textit{aggressive}, is generally difficult
and interpreting the noisy confidence estimate of the
intent prediction is prone to errors.
Hence, in this work we propose an initial step towards a general shared autonomy approach
where arbitration strategies can be learned from user interaction.
We do this by computing an optimal blending parameter in hindsight,
and train an RNN by supervised learning to predict optimal arbitration.
The system is trained by data aggregation which is beneficial for two reasons.
First, it reduces state distribution shift \cite{ross2010efficient}
and second, it allows the system to adapt to users online.

Our framework is based on a differentiable graph model of the policy,
see Figure \ref{fig:architecture},
that integrates motion generation, user intent inference and arbitration.
The motion generation module
is trained to mimic a trajectory optimizer.
The intent inference module is trained to predict
goal distributions using data collected in a direct control phase.
In our experiment the arbitration function is either hand-crafted or trained
through user interaction.

To assess the efficacy of the approach, we define a teleoperation task
where a user controls a virtual gripper using velocity commands using hand motion $v_{hand}$.
We perform a pilot user experiment and compare completion times
on two different tasks: with and without obstacles.

\section{Related Work}

\subsection{Managing autonomy in teleoperation}

Any robot to be teleoperated presents a certain level of autonomy.
Managing this autonomy is crucial in the design of a teleoperation system
\cite{Ferrell:67, Sheridan:92}.
Bellow we present the three most relevant ways of managing the autonomy in 
supervisory control \cite{Goodrich:13}.

\subsubsection{Direct control}

The operator manually controls the robot; no autonomy is involved on the robot side. 
The robot is controlled in a master-slave interaction. 
An example of direct-control is the control of each DoF of a manipulator using a joystick.

\subsubsection{Traded control}

The operator initiates a sub-task or behavior for the robot and the robot performs the sub-task autonomously while the operator monitors the robot.
In the DARPA Robotics Challenge (DRC), most teams implemented this approach
by having the operator specify the object to be manipulated or goal location to walk to.
The robot then performed the task using motion planning algorithms \cite{phillips2016autonomy, phillips2014toward}.
This approach is particularly relevant to low bandwidth teleoperation.

\subsubsection{Shared control}

The operator and the robot simultaneously control the robot.
To solve the control blending strategy,
approaches include using model-free deep reinforcement learning
\cite{reddy2018shared} in discrete action spaces,
or formulating the problemas a Partially Observable Markov Decision Process (POMDP)
and solving the problem using hindsight optimization \cite{Javdani:2018bt}.
However, this approach results in mixed user's satisfaction compared to policy blending
\cite{dragan2013policy} despite improved task performance.
Our work is based on a shared autonomy framework,
where we learn the arbitration (i.e., blending strategy).
In our experiments we compare our work with a direct control approach.

\subsection{Imitation Learning}

Inverse Reinforcement Learning \cite{Ng:2000tv,Ziebart:08, Mainprice:16b}
can identify the optimal criterion used when taking sequential decisions
and is often utilized in the inner loop of shared autonomy systems.
However, it solves an ill-posed problem that
suffers from reward and policy ambiguity.
In this work we rather apply a simple behavior cloning
approach for a goal that is predicted using a Recurrent Neural Network (RNN),
which provides confidence scores.

Our approach is based on similar ideas 
developed for imitation learning, namely data aggregation \cite{ross2010efficient},
which consists of iterative supervised training and user interaction 
to learn an arbitration function without covariate shift.
This is necessary since the user behavior depends on the
arbitration function itself, introducing a learning feedback loop.

\section{Method}

We created a simulated 2-dimensional pick-and-place environment
as shown in the top-left image in Figure \ref{fig:architecture}. The robot is a simple freeflying 2DoF gripper (depicted in blue). Let \(x \in X\) the continuous robot state (position, velocity) and \(a_u \in A_u\), \(a_r \in A_r\), \(a_s \in A_s\) be the continuous actions (velocity) of the user, autonomous agent, and the shared autonomy agent. The square objects represent a discrete set of possible goals \(g \in G\). The task is to control the gripper using hand motion to pick the gray object and retrieve it back to the green goal position.

\begin{figure}[t]
  \centering
  \includegraphics[clip,trim={3.5cm 3.5cm 3.5cm 3.5cm},scale=0.35]{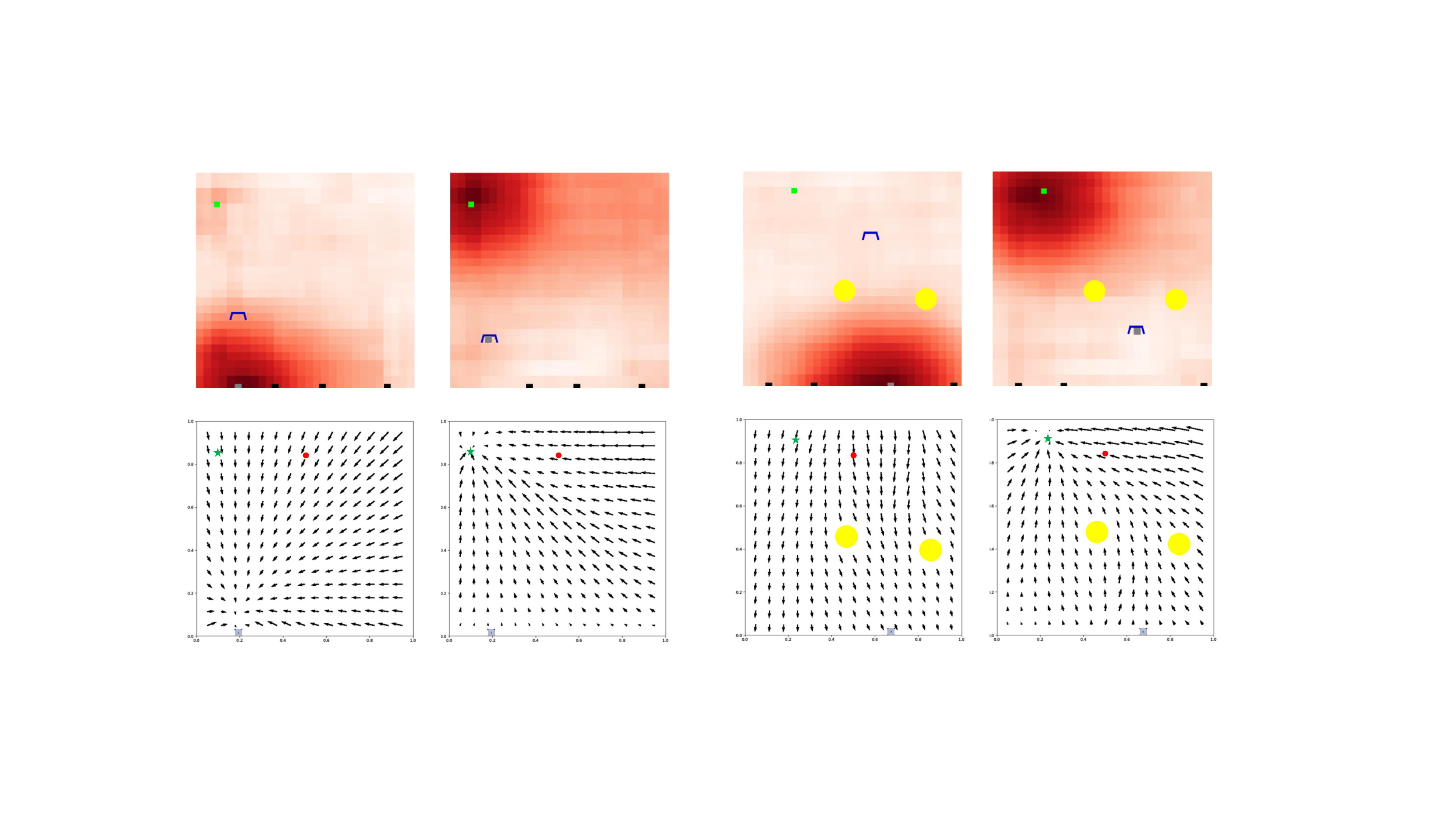}
  \caption{Top: predicted intent 28x28 likelihood heatmaps.  
  Bottom: Robot policy for pick-and-place tasks in an environment
  with and without obstacles (resp., two left and two right columns)}
  \label{fig:heat_and_quiv}
  \vspace{-.5cm}
\end{figure}
\subsection{User intention inference}
The autonomous agent must first predict the user intentions before assisting the user action.   
We predict the likelihood of intended goal position given observed hand movement as a 28x28 discrete heatmap grid over the environment (see Figure \ref{fig:heat_and_quiv}). 
We train a Long Short-Term memory (LSTM) network, followed by a convolution transpose layer to predict value for each grid cell.
The module is trained on direct control data from 19 participants (approx. 1400 episodes). 

Previous works have proposed different ways to compute confidence values for the goal and defined a fixed relationship between confidence and arbitration
\cite{dragan2013policy, gopinath2016human, schultz2017goal}.
Here, we directly use the scores \(c \in C\) from the grid corresponding to each object $g_{i}$. 
The predicted goal object $g^*=\argmin_o\sum_{n=1}^{k}c_n$ 
is the object with the maximum sum of scores over a time sequence $k$.  

\begin{figure*}[t]
  \centering
  \includegraphics[width=.21\linewidth]{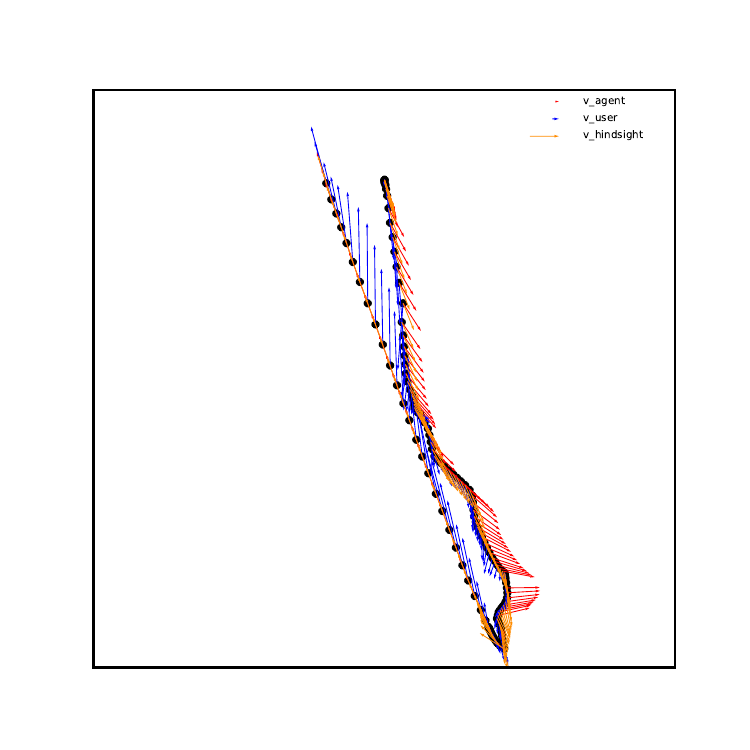}
  \includegraphics[width=.28\linewidth]{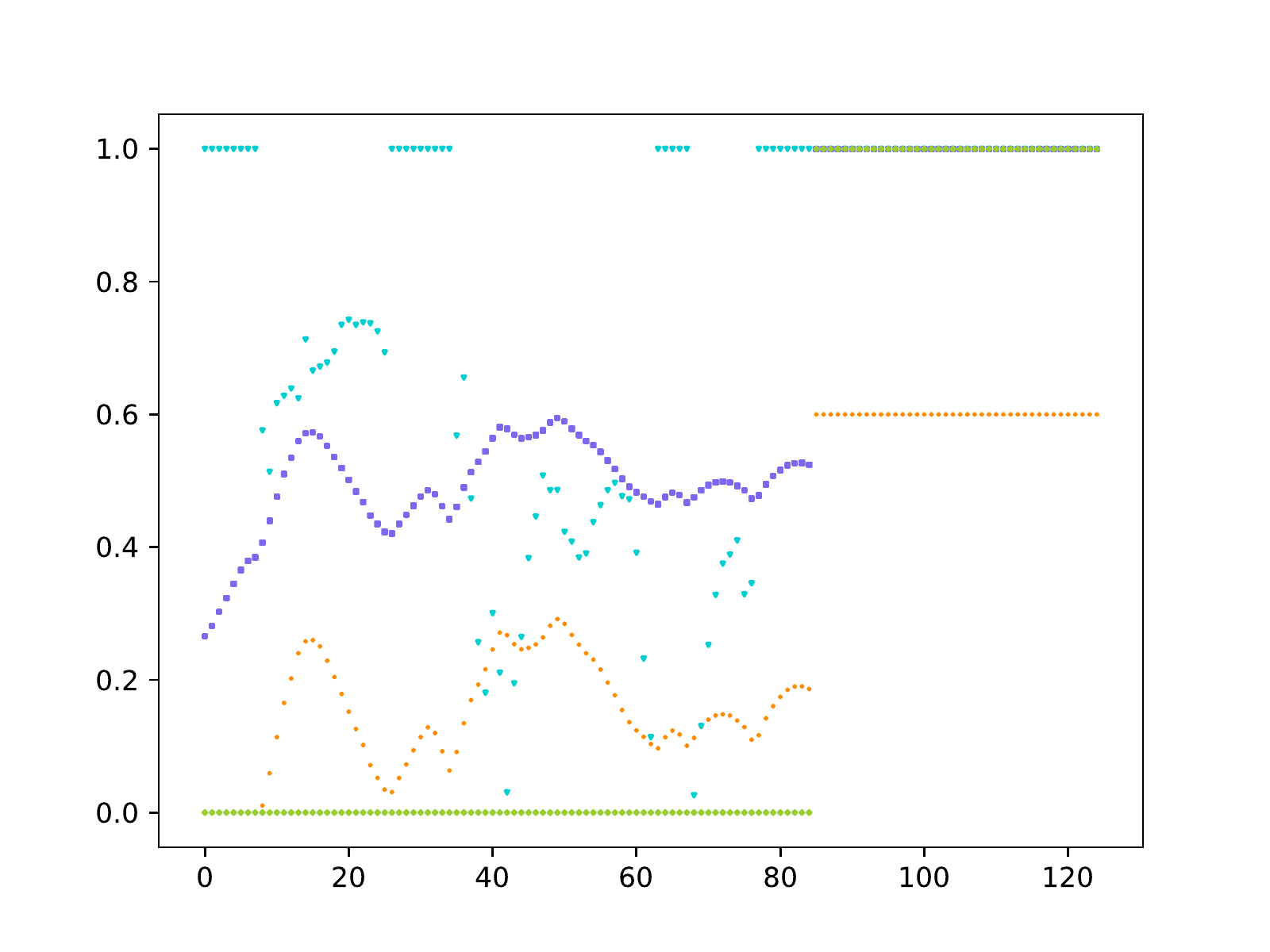}
  \includegraphics[width=.21\linewidth]{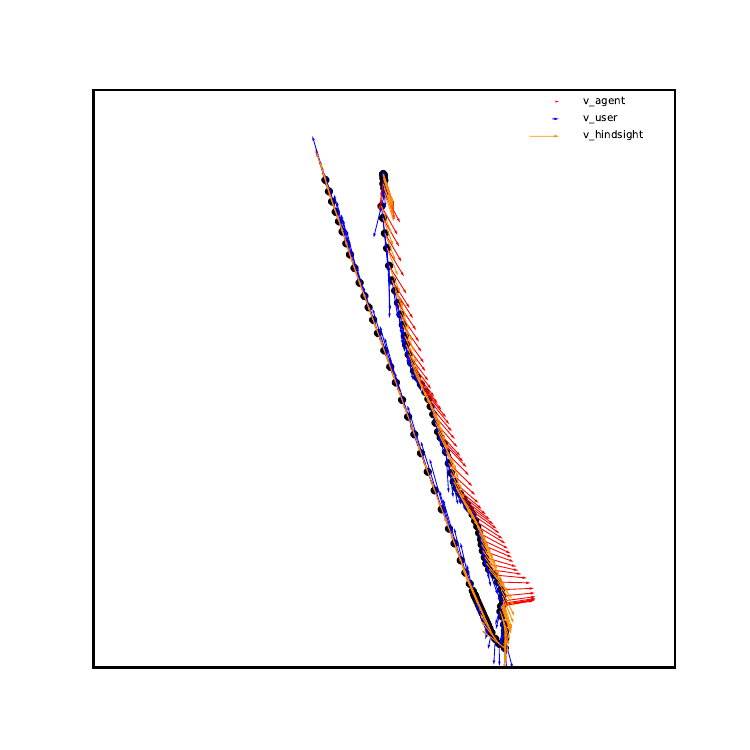}
  \includegraphics[width=.28\linewidth]{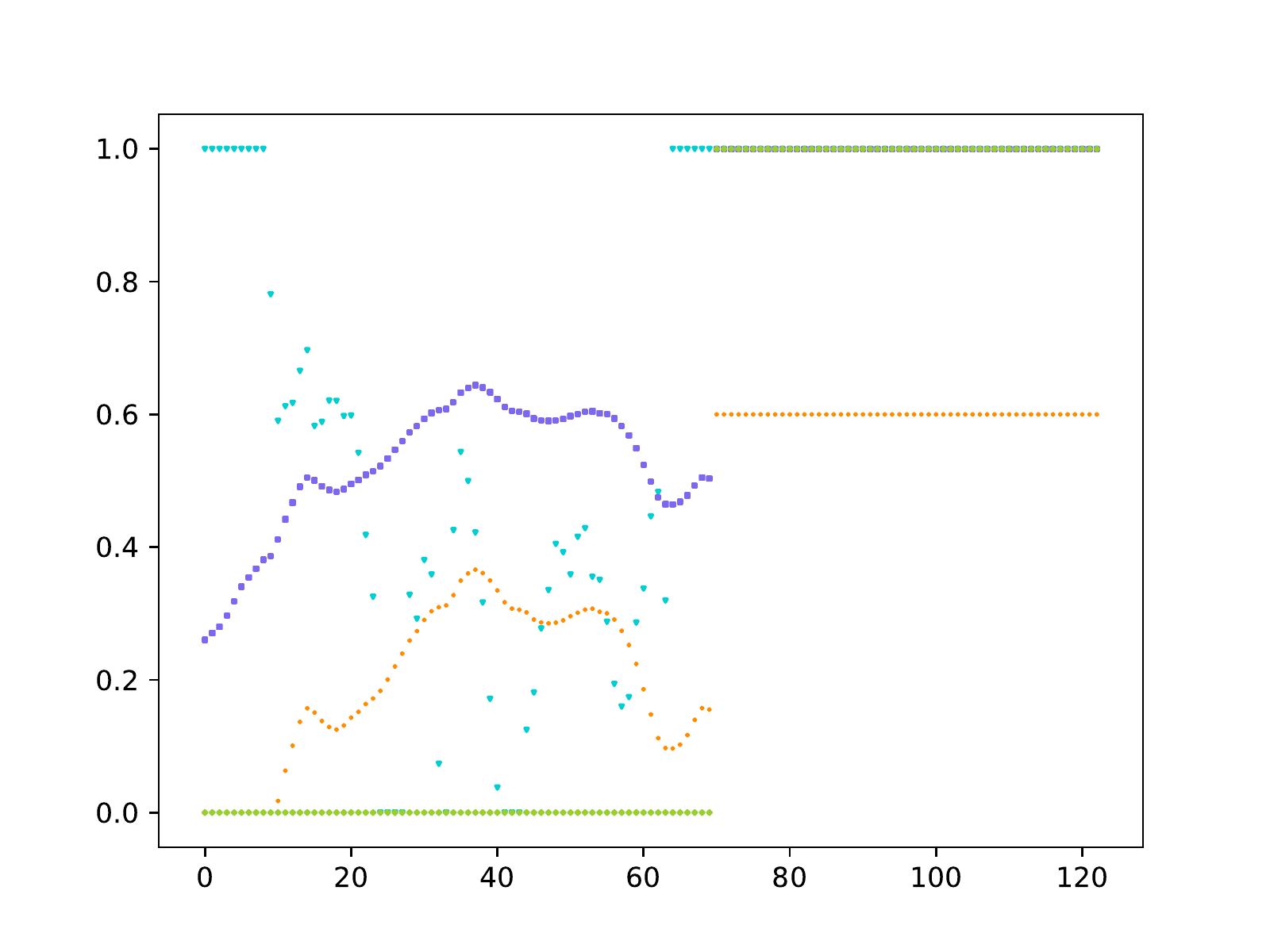}
  
  \caption{Prediction and assistance for the wrong goal. 
  Left (1 and 2): alpha trained with 30 episodes. Right (3 and 4): alpha trained with 120 episodes.
  Trajectory of the gripper along with commands (1 and 3). 
  Alpha (cyan), grabbed state(light green), confidence (purple) and baseline timid arbitration (orange) \cite{dragan2013policy},
  as a function of time steps (2 and 4). In the later run, the arbitration module learns
  to give control authority back in the first phase of the grasping motion to compensate
  for low confidence values, resulting in a smoother trajectory. Note $v_{\text{agent}}$ in red
 always point to the right in the direction of a wrongly estimated intent. }
   \label{fig:trajs_and_alphas}
\end{figure*}

\subsection{Motion generation}
Given a predicted goal, the autonomous agent generates an optimal action at each timestep. 
We train a Multi-Layer Perceptron (MLP) network to mimic trajectories generated from a Gauss-Newton trajectory optimizer \cite{mainprice2016warping}.
This step is similar to the first step of guided policy search (i.e., supervised learning based on optimal control data of pick-and-place
motion given the positions of the object to grasp and the position to place). 

The motion generation policy $\pi_r$ is regressed to the whole state space. Thus, given the state of the robot and the predicted goal object $g^*$, we can infer the optimal action $v_{robot}$ at each timestep.
The network is trained using 3000 successfully optimized trajectories. For the environment with obstacles, we use 24K optimized trajectories.

\subsection{Arbitration learning by hindsight data aggregation}
Different arbitration functions can lead to different assistance behavior of the shared autonomy agent. It is difficult to define a fixed arbitration function considering how well the the intent is predicted and the user's preference to assistance levels. 
When the goal predicts and assists for the wrong goal, the user should keep control authority and guide the robot towards the intended goal. 
We consider the learning problem of an arbitration function that takes into account the failure in intent prediction as well as the user's adaptability towards assistance. 


We record teleoperation (i.e., direct and shared control) trajectories and calculate the optimal action in hindsight at each timestep using the motion generation policy, which we call hindsight action \(a_h\). The aim is to find an arbitration value $\alpha$ such that \(a_s\) matches \(a_h\).
A commonly used blending function is:
\begin{equation}
  \label{equ:linear_arbitration}
  a_s=\alpha a_r+(1-\alpha)a_u
\end{equation}

However, here we consider a slightly different control policy where the user keeps
authority over velocity and is solely assisted with the gripper direction.
Hence, arbitration is simply an angle $\theta$, between user direction $a_u$ 
and autonomy direction $a_r$.
Therefore, we use the following non-linear blending function:

\begin{equation}
  \label{equ:geometric_arbitration}
    a_s  = R (\theta) a_u, \\
\end{equation}
where $R$ is a rotation matrix and $\theta  =  \alpha \theta_{ur}$. 
That is, $\alpha$ is the fraction of the angle $\theta_{ur}$. 
Which, leads to the following equation for $\alpha$:

\begin{equation}
  \label{equ:arbitration_calculation}
  \alpha=
    \begin{cases}
      0, & \text{if } \theta_{uh} = \theta_{rh}-\theta_{ur} \\
      1, & \text{else if } \theta_{uh} = \theta_{ur}+\theta_{rh} \\
      \frac{\cos ^{ - 1}(\hat{a_u}\cdot \hat{a_h})}{\cos ^{ - 1}(\hat{a_u}\cdot \hat{a_r})}, & \text{otherwise } 
    \end{cases}
\end{equation}

where $\theta_{uh}$, $\theta_{rh}$, $\theta_{ur}$ denote the angles between \(a_u\), \(a_h\) and \(a_r\), \(a_h\), and \(a_u\), \(a_r\) and $\hat{a_u}$, $\hat{a_h}$ denote the unit vectors of corresponding actions.
That is, $\alpha$ is the ratio between angles $\theta_{uh}$ and $\theta_{ur}$ when \(a_h\) lies between vectors \(a_u\) and \(a_r\), we clip $\alpha$ when \(a_h\) is outside of this range.

We use a LSTM network followed by dense layers to predict a value and minimize the mean squared error between predicted and target alpha values. The network is pretrained using direct control data. Training the network only with direct control data is unlikely to generate a robust arbitration function as $\alpha$ becomes close to 1 when the user intent inference is correct. 

Hence, we iteratively collect sets of data in shared control mode using arbitration predicted by the network and aggregate into the dataset. The model is then retrained using the aggregated dataset. This online training allows to capture the disagreement between \(a_u\) and \(a_s\) such that the user can have more authority when the shared autonomy agent's policy differs from the user's.

\section{Preliminary Experiment and Results}

We first performed user experiment with eight subjects and asked to execute the pick-and-place (using a Leap Motion hand gesture sensor).
The 2D environment is displayed in a tilted angle to simulated a perspective view from the camera mounted on a robot. 

Note that in this experiment, we do not perform arbitration learning but calculate a confidence value from intent prediction and map it to an arbitration function similar to \cite{dragan2013policy}.

\begin{figure}[t]
  \centering
  \includegraphics[height=.5\linewidth]{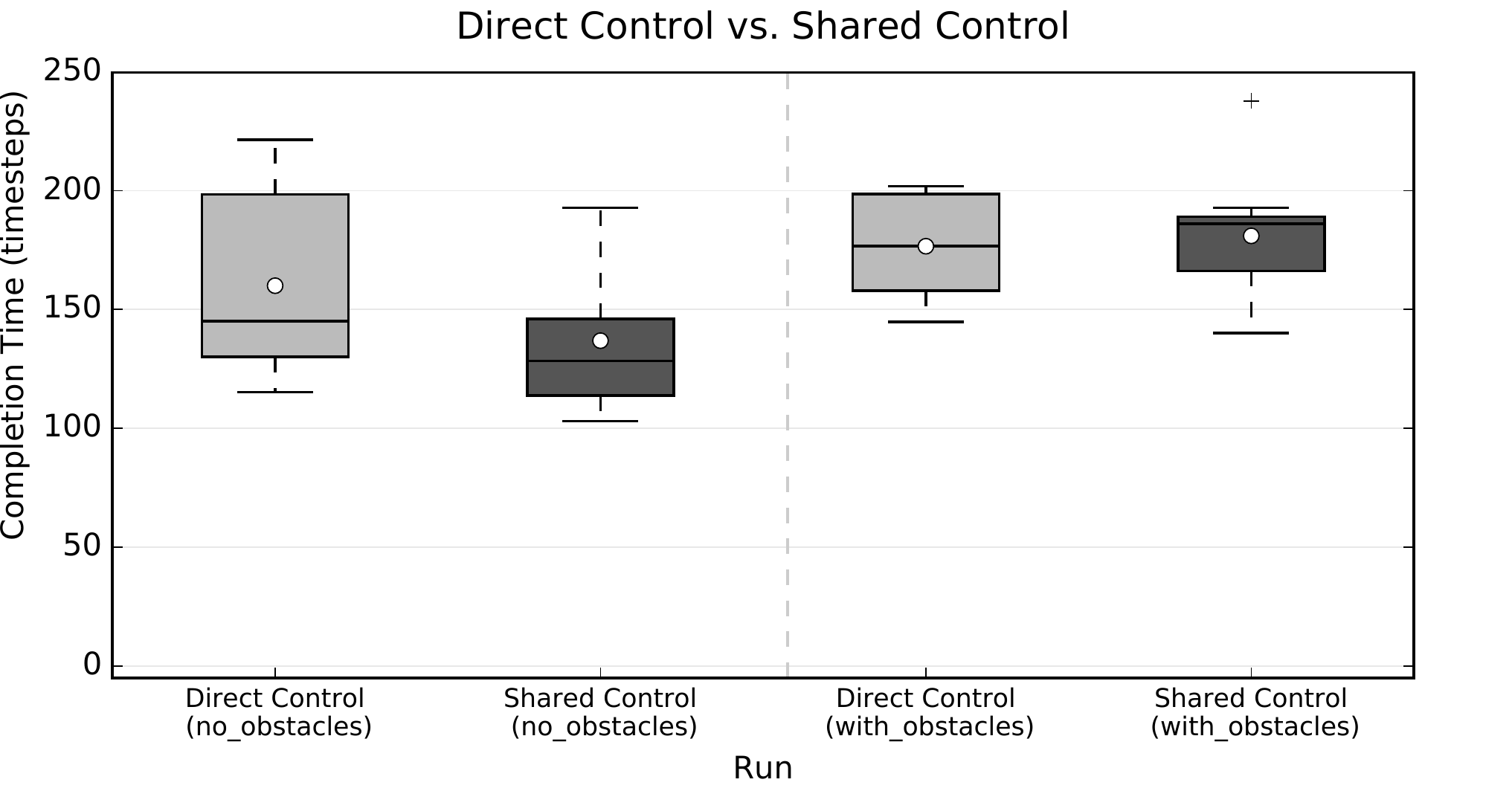}
  \caption{Comparing the average completion time of across all episodes for direct control and shared control,
  with and without obstacles.}
  \label{fig:all_boxplot}
  \vspace{-.5cm}
\end{figure}

Subjects performed the experiment for two different modes, (i.e. direct and shared control),
for a sequence of 12 episodes twice in random order. 
Subjects were allowed to practice each mode for a very short time before the test.
The experiment was repeated in both with and without obstacles present environments.
Figure \ref{fig:all_boxplot} shows the result of average completion times for both control modes and environments.

From the environment without obstacles, it is shown that shared autonomy improved the performance of teleoperation. 
However, in the environment with obstacles,
the performance using the shared autonomy is slightly lower than with direct control. This is caused by the incorrect user intent prediction, which was trained in an environment without obstacles. The curved trajectories made by the participants often lead to wrong goal predictions. As a consequence subjects had to "fight" the controller to make the
gripper move towards the subjects' true intention. This could be alleviated by carefully characterizing confidence and arbitration function, however, it shows the need for an adaptable arbitration that takes into account the inaccuracy of the prediction and user's behaviors.

\begin{figure}[t]
  \centering
  \includegraphics[height=.5\linewidth]{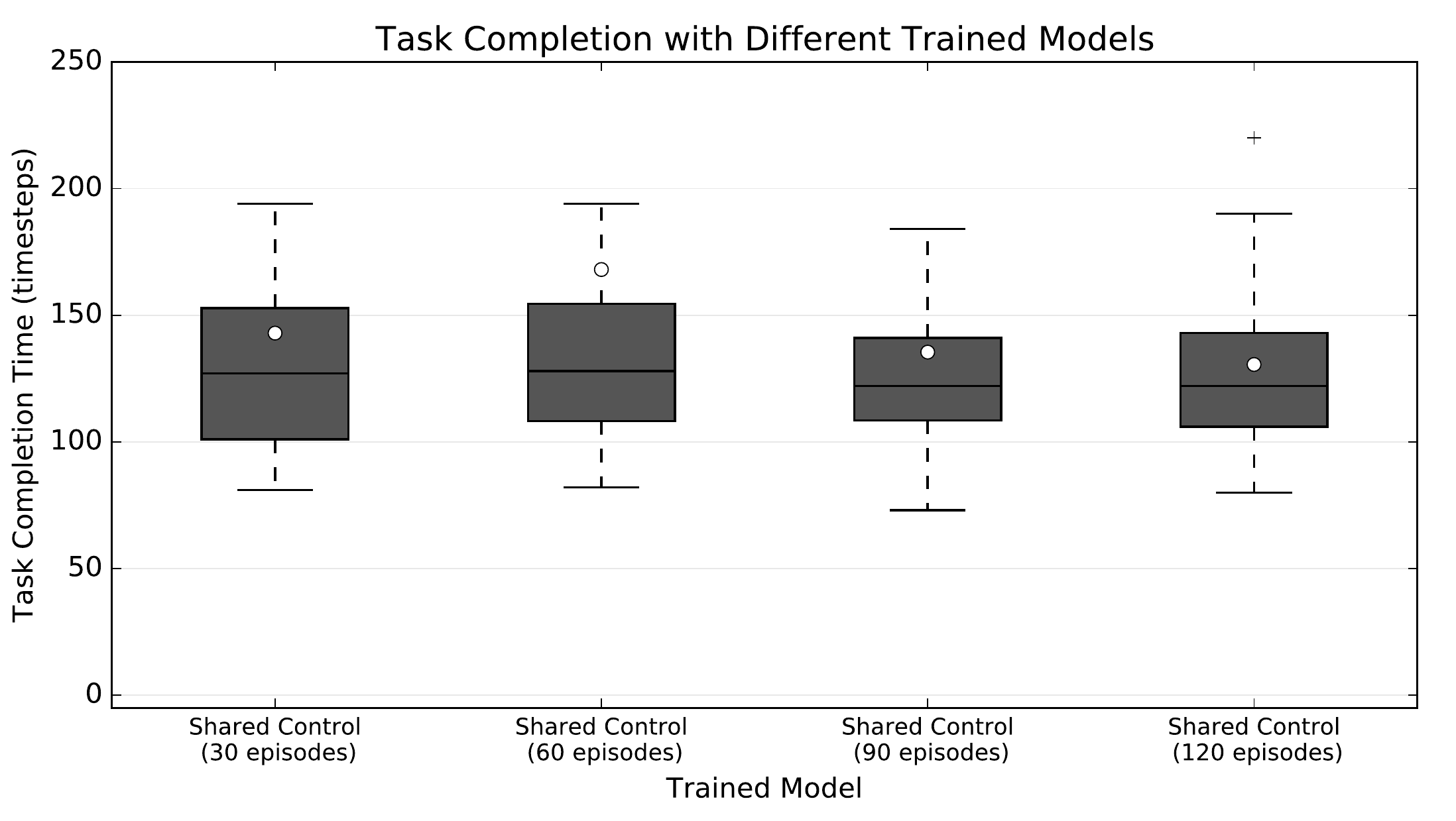}
  \caption{Comparing the average completion time for each mode that is trained with different aggregation dataset size}
  \label{fig:task_completion}
\end{figure}

Next, we show preliminary results using our arbitration training model. We show pick-and-place execution performance in the 2D environment with no obstacles using the model trained with different sizes of the teleoperation dataset. Five subjects performed the experiment using four arbitration function models 10 times each in a random order. Each model shown on the x-axis in Figure \ref{fig:task_completion} indicates the arbitration model trained with the aggregated dataset. Each model is initialized using the parameters of the trained model on its left. 
The first model is trained using 30 shared control episodes using the model pretrained with 100 direct control trajectories. The second model is trained using 60 shared control episodes using the model initialized with parameters of the first model. Using each model we collect 30 new trajectories and the data is aggregated in the dataset. In the same way, the third mode denotes the model trained with 90 episodes (30 new + 60 in dataset) with the model parameters restored from the second mode.

The overall task performance using trained arbitration showed better performance compared to direct control performance shown in Figure \ref{fig:all_boxplot}. Also, it was similar to that using the the linear blending arbitration function. Between different trained model settings we could see a performance increasing, this shows that the more episodes are aggregated to the dataset
the more the arbitration becomes robust.
This is also shown in Figure \ref{fig:trajs_and_alphas} where the shared control trajectory using alpha trained with 120 episodes is smoother than the shared control trajectory using alpha trained with 30 episodes, which means that the user gained more authority in shared control when the user intent inference model predicts the wrong goal. 

\begin{figure}[h]
  \centering
  \includegraphics[clip,trim={1.5cm 1.5cm 1.5cm 1.5cm},scale=0.37]{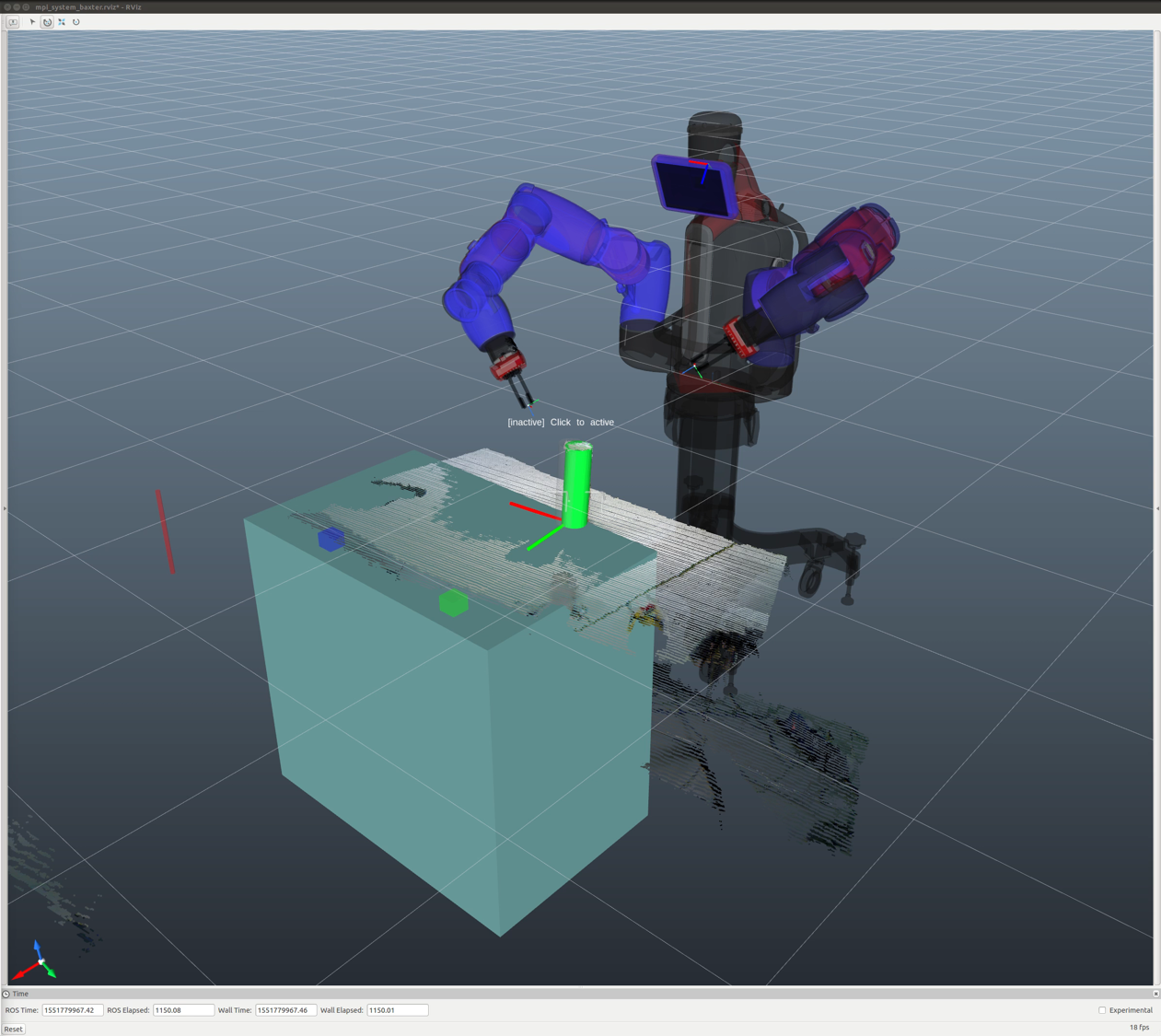}
  \caption{Baxter setup using RGBD camera
  with object and arm tracking. Baxter has with two 7DoF arms,
  which make the problem more challenging.}
  \label{fig:baxter_setup}
\end{figure}
\vspace{-.2cm}



\section{Conclusion and Future Work} 
\label{sec:conclusion}

In this work we introduced a differentiable model and training method 
to learn arbitration functions and show preliminary
results using the proposed framework.
We performed a pilot user experiment where it was shown that we can learn an arbitration to substitute hand-crafted arbitration functions. 

Future work includes extending user experiments and showing improvement in qualitative as well as quantitative results. In addition, we plan to test our framework using a robot setup based on \cite{kappler2018real} as shown in Figure \ref{fig:baxter_setup} as we believe that we can transfer the 2D simulation policy to the planar teleoperation of the end-effector in task space control. Thus, our goal is to express a formalism for shared control that is able to be trained end-to-end for a robust shared autonomy system that can adapt to each user and provide seamless blending between user and robot.

\section*{ACKNOWLEDGMENT}
This work is partially funded by the research alliance ``System Mensch''.
The authors thank the International Max Planck Research School for Intelligent Systems (IMPRS-IS)
for supporting Yoojin Oh.

\clearpage
\bibliographystyle{plainnat}
\bibliography{references}
\balance

\end{document}